\documentclass[final]{cvpr}

\usepackage{times}
\usepackage{epsfig}
\usepackage{graphicx}
\usepackage{amsmath}
\usepackage{amssymb}
\usepackage{bm}
\usepackage{color}
\definecolor{mypink}{RGB}{255,135,180}
\definecolor{myyellow}{RGB}{255,200,0}
\definecolor{mygreen}{RGB}{0,100,0}
\definecolor{mygold}{RGB}{255,185,0}
\definecolor{mymaroon}{RGB}{128,0,0}
\definecolor{myaquamarine}{RGB}{120,200,190}
\definecolor{myturquoise}{RGB}{64,224,208}
\definecolor{mymagenta}{RGB}{255,0,255}
\definecolor{myviolet}{RGB}{238,130,238}

\usepackage{enumitem}
\setenumerate[1]{itemsep=0pt,partopsep=0pt,parsep=\parskip,topsep=2pt}
\setitemize[1]{itemsep=0pt,partopsep=0pt,parsep=\parskip,topsep=2pt}
\setdescription{itemsep=0pt,partopsep=0pt,parsep=\parskip,topsep=2pt}

\makeatletter
\def\thickhline{%
  \noalign{\ifnum0=`}\fi\hrule \@height \thickarrayrulewidth \futurelet
   \reserved@a\@xthickhline}
\def\@xthickhline{\ifx\reserved@a\thickhline
               \vskip\doublerulesep
               \vskip-\thickarrayrulewidth
             \fi
      \ifnum0=`{\fi}}
\makeatother

\newlength{\thickarrayrulewidth}
\usepackage[pagebackref=true,breaklinks=true,colorlinks,bookmarks=false]{hyperref}

\def\cvprPaperID{2811} 
\def\confYear{CVPR 2021}

\def\abbrapproach{ASAL}

\begin{document}

\title{Action Shuffle Alternating Learning for Unsupervised Action Segmentation}

\author{Jun Li\\
Oregon State University\\
{\tt\small liju2@oregonstate.edu}
\and
Sinisa Todorovic\\
Oregon State University\\
{\tt\small sinisa@oregonstate.edu}
}

\maketitle
\thispagestyle{empty}

\begin{abstract}
   This paper addresses unsupervised action segmentation. Prior work captures the frame-level temporal structure of videos by a feature embedding that encodes time locations of frames in the video. We advance prior work with a new self-supervised learning (SSL) of a feature embedding that accounts for both frame- and action-level structure of videos. Our SSL trains an RNN to recognize positive and negative action sequences, and the RNN's hidden layer is taken as our new action-level feature embedding. The positive and negative sequences consist of action segments sampled from videos, where in the former the sampled action segments respect their time ordering in the video, and in the latter they are shuffled. As supervision of actions is not available and our SSL requires access to action segments, we specify an HMM that explicitly models action lengths, and infer a MAP action segmentation with the Viterbi algorithm. The resulting action segmentation is used as pseudo-ground truth for estimating our action-level feature embedding and updating the HMM. We alternate the above steps within the Generalized EM framework, which ensures convergence.  Our evaluation on the Breakfast, YouTube Instructions, and 50Salads datasets gives superior results to those of the state of the art. 
\end{abstract}

\section{Introduction}
\label{sec:Intro}

\begin{figure*}[t]
\centering
\begin{minipage}{0.35\linewidth}
\centering
\includegraphics[width=0.99\linewidth]{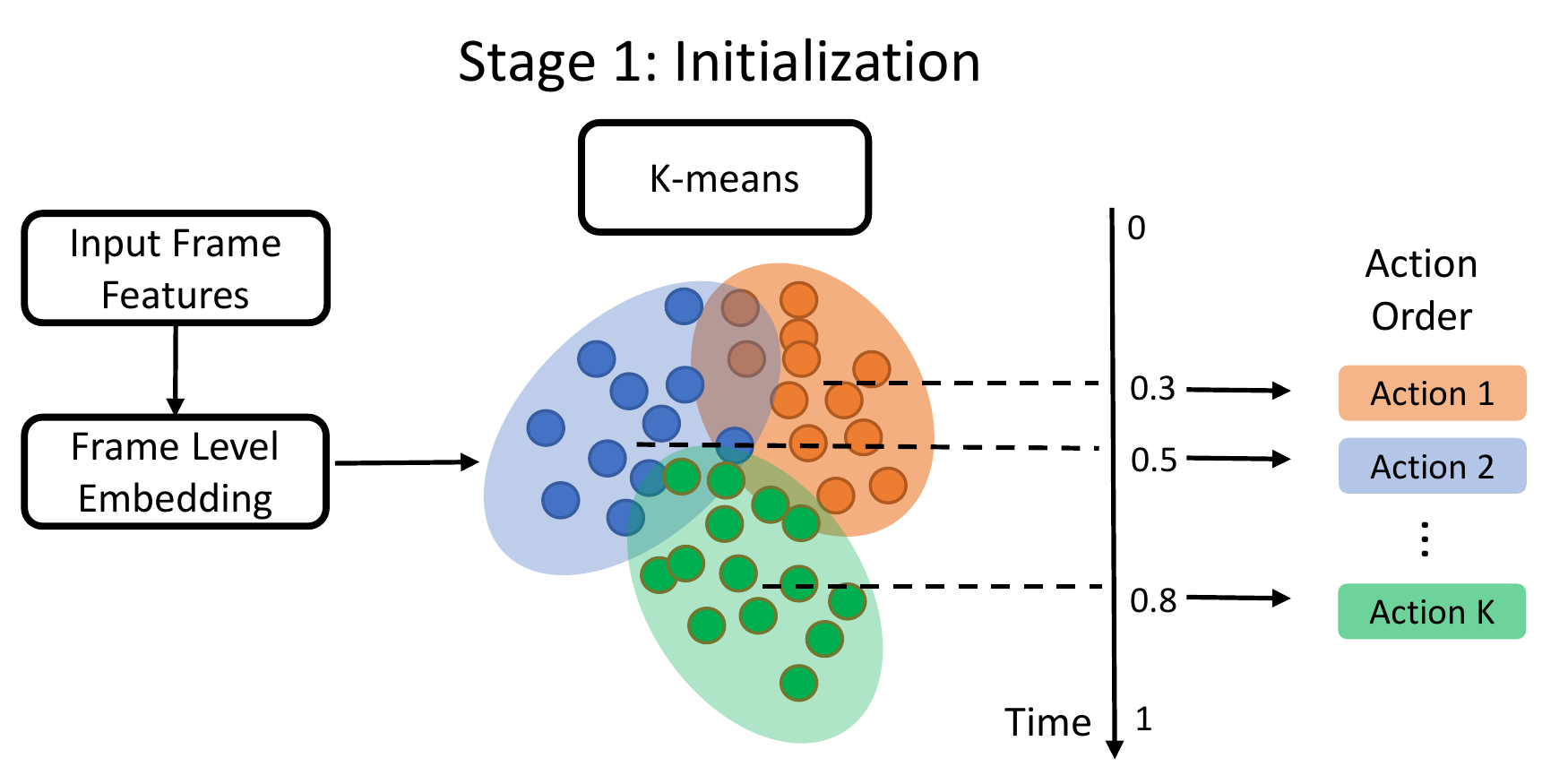}

(a) 
\end{minipage}\hfill%
\begin{minipage}{0.3\linewidth}
\centering
\includegraphics[width=0.99\linewidth]{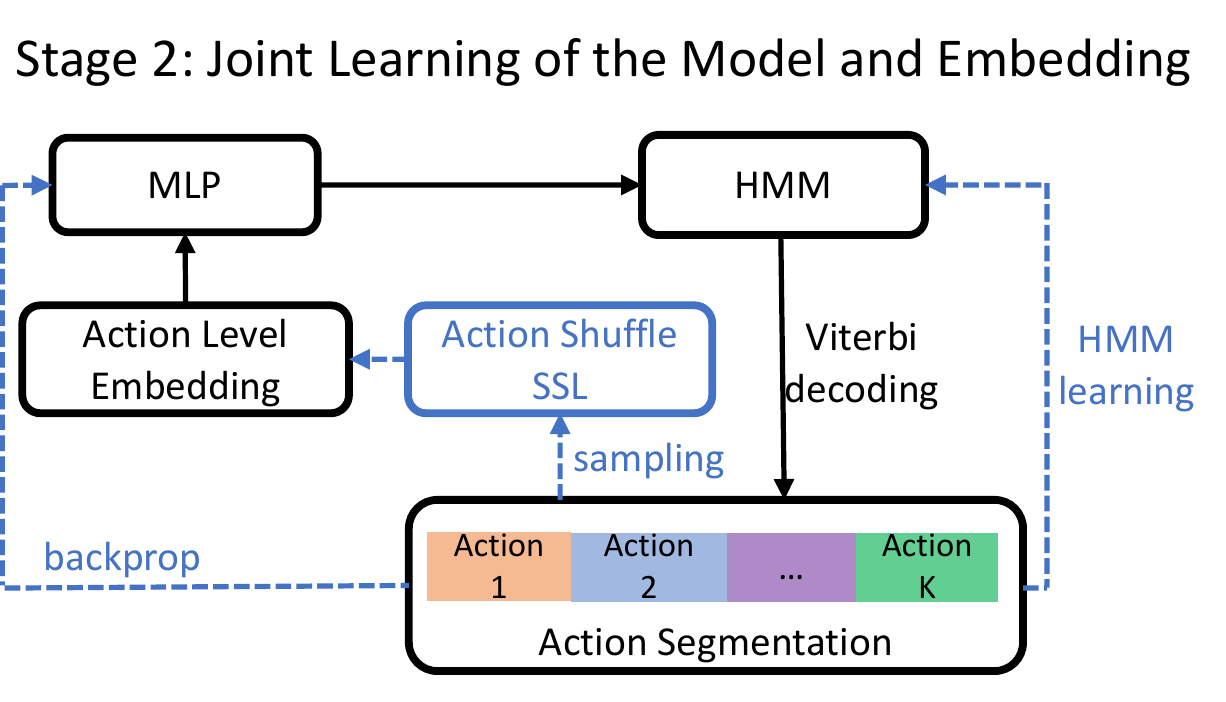}
   
(b)
\end{minipage}\hfill%
\begin{minipage}{0.3\linewidth}
\centering
\includegraphics[width=0.99\linewidth]{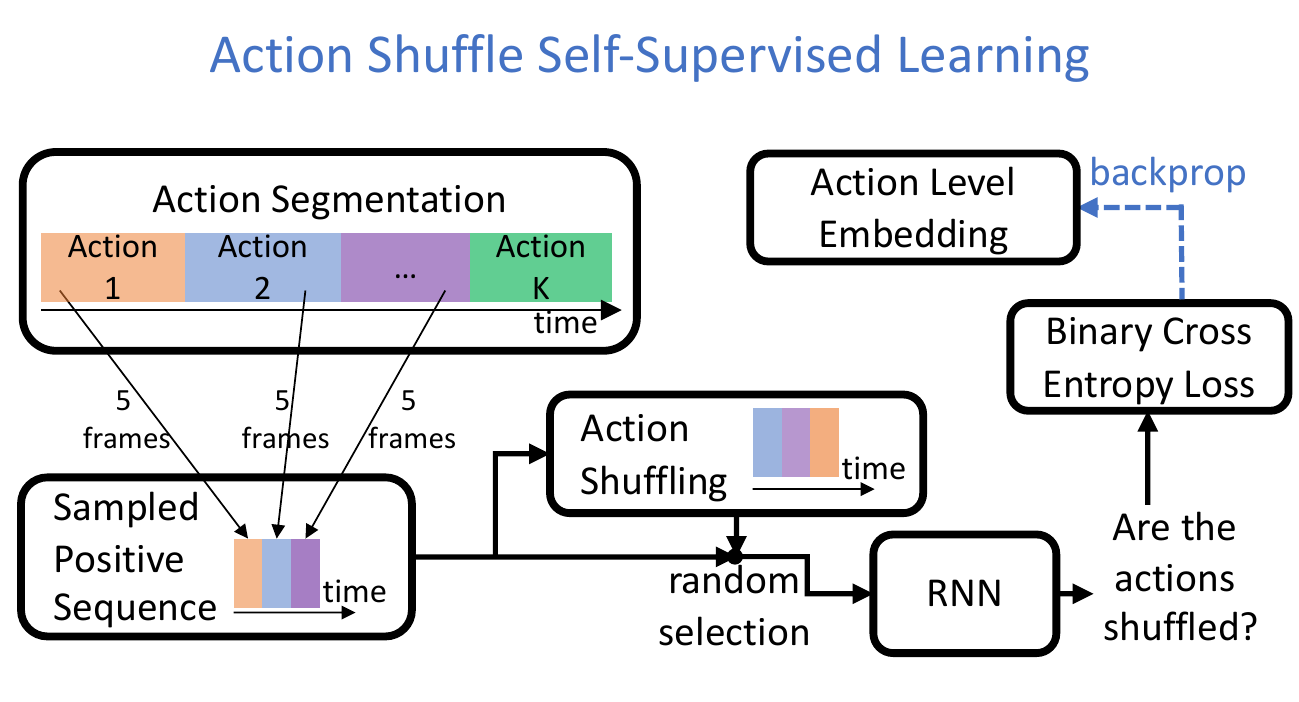}
   
(c) 
\end{minipage}
\caption{An overview of our unsupervised learning.  (a) Initial clustering of frame features for identifying latent actions, and their temporal ordering; the frame-level feature embedding is from \cite{kukleva2019unsupervised}. (b) The iterative joint training of HMM, MLP, and Action Shuffle SSL for inferring action segmentation within the generalized EM framework. (c) SSL for learning our action-level temporal feature embedding. The data manipulation samples positive and negative sequences, where the former respect temporal ordering of actions in the predicted action segmentation and the latter shuffle the ordering of actions.}
\label{fig:Overview}
\end{figure*}

This paper is about unsupervised action segmentation, where the goal is to localize salient latent actions in untrimmed videos. The actions are salient as differences in their features allow for segmentation, and the actions are latent as they may not have a distinct semantic meaning, since no supervision about the actions is available. The ground truth can be used only for evaluation. This is a long-standing vision problem with a wide range of applications. It can be used for mapping a long video to a significantly shorter sequence of action segments, and thus for facilitating and reducing complexity of subsequent video interpretation.  It can also be used in applications where manual video annotation is prohibitively expensive or not reliable. 

In this paper, we focus on a particular setting studied in recent work \cite{kukleva2019unsupervised},  where all videos show the same activity (e.g., a cooking activity) which can be decomposed into a temporal sequence of simpler actions (e.g., cooking includes cutting, mixing, peeling). While the activity exhibits variations across videos, they are mostly manifested in varying lengths and features of each action of the activity, whereas variations in the total number and temporal ordering of actions are limited by the very nature of the activity (e.g., cooking usually requires a certain order of actions).

For such a setting, related work \cite{kukleva2019unsupervised} makes the following restrictive assumptions that every action appears only once, and all actions always occur and follow the same temporal ordering in all videos, referred to as {\em fixed transcript}. Based on these assumptions, they learn a temporal feature embedding by training a regression model to predict a temporal position of every frame in the video, where the frame positions are normalized to the video length. Then, they use the K-means for clustering these embedded features of all video frames, and interpret the resulting clusters as representing the latent actions. After computing a likelihood for every cluster, they run the standard Viterbi algorithm on every video for localizing the latent actions.

We make {\em three contributions}, as illustrated in Fig.~\ref{fig:Overview}. First, we relax the above assumption about the fixed temporal ordering of actions in videos. We specify a Hidden Markov Model (HMM), and thus infer a MAP ordering of actions, instead of the fixed transcript. Unlike \cite{kukleva2019unsupervised}, our HMM explicitly models the varying lengths of latent actions, and thus constrains implausible solutions in the domain (e.g., chopping cannot take a few frames). Also, our HMM uses a multilayer perceptron (MLP) for estimating the likelihood of the frame labeling with the latent actions. 

Second, we specify a new self-supervised learning (SSL) for {\em action-level temporal feature embedding}. As other SSL approaches \cite{kukleva2019unsupervised, misra2016shuffle, JenniECCV20, LeeHS017, Ommer18, Kim_Cho_Kweon_2019, EpsteinCV20, BenaimELMFRID20,Yao_2020_CVPR}, ours exploits the temporal structure of videos, where the structure we mean that videos are sequences of actions with small variations in action ordering. However, the cited references typically focus on capturing the temporal structure at the frame level (e.g., by shuffling or permuting frames \cite{misra2016shuffle, LeeHS017}, encoding frames' positions in video \cite{kukleva2019unsupervised}) or the entire video level  (e.g., by manipulating the playback speed \cite{EpsteinCV20, BenaimELMFRID20,Yao_2020_CVPR}). In contrast, we seek to learn a feature embedding that would account for the correct ordering of actions along the video. 

\begin{figure}
\centering
\includegraphics[width=\linewidth]{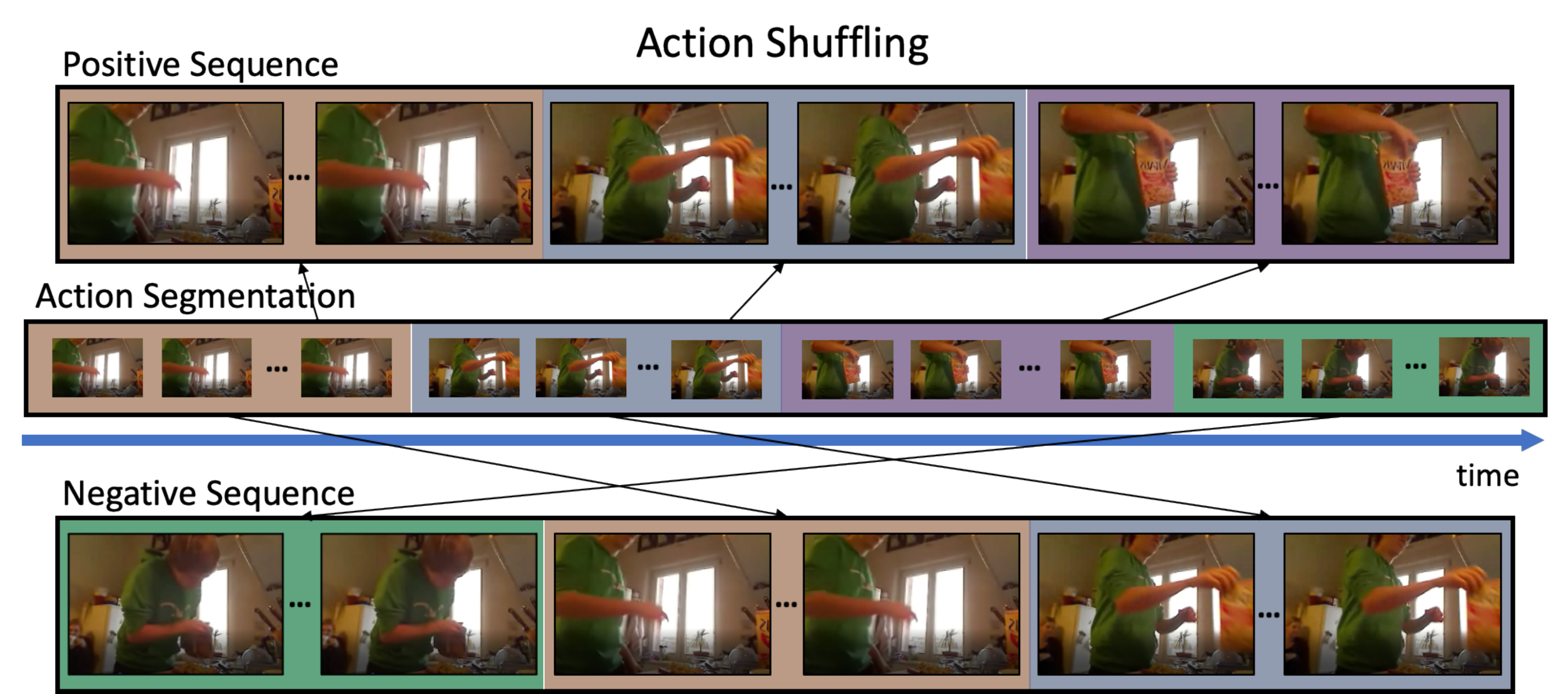}
\caption{Action shuffling: Both positive and negative examples are generated as sequences of 3 action segments randomly sampled from a video, where the former respects the action ordering in the video, and the latter shuffles the ordering.}
\label{fig:ActionShuffling}
\end{figure}

As illustrated in Fig.~\ref{fig:Overview} and  Fig.~\ref{fig:ActionShuffling}, our SSL is based on recognizing positive and negative sequences of actions, both generated as sequences of action segments randomly sampled from a video. The key difference between the positive and negative sequences is that in the former the actions are laid out in the same temporal ordering as in the video, and in the latter the actions are shuffled resulting in an incorrect ordering. As shown in Fig.~\ref{fig:Overview}, our SSL trains an RNN on these positive and negative sequences, and minimization of the incurred binary cross-entropy loss produces our action-level temporal embedding of frame features. 

As our third contribution, we formulate a joint training of HMM, MLP, and Action Shuffle SSL within the Generalized EM framework \cite{tanner2012tools}. This extends prior work (e.g. \cite{kukleva2019unsupervised, SaxenaCVPR15})  where different components of their approaches are usually learned independently. Note that our Action Shuffle SSL requires access to action segments for generating the positive and negative examples. Action segmentation, in turn, requires the HMM inference. After the HMM inference, we can update the action-level feature embedding through the Action Shuffle SSL, as well as update the HMM parameters. As these updates will change both features and HMM, it seems reasonable to run the HMM inference again. We integrate all these steps within the generalized EM framework. A convergence guarantee of our joint training follows from the generalized EM algorithm \cite{tanner2012tools}.

Fig.~\ref{fig:Overview} shows an overview of our unsupervised learning that consists of two stages. The initial stage follows \cite{kukleva2019unsupervised}. Given frame features from all videos, we first compute the frame-level feature embedding of \cite{kukleva2019unsupervised}, and then run the K-means for identifying labels of the latent actions. As shown in Fig.~\ref{fig:Overview}a, for every cluster, we compute the mean of frames' normalized positions in videos, then, sort the clusters by their means in the ascending order, and finally take the sorted cluster indices as labels of the corresponding latent actions. This ascending order of action labels is taken as a likely transcript, and used in the HMM inference for constraining the Viterbi algorithm to respect the transcript. In the second stage, we perform the generalized EM algorithm for iteratively updating the HMM, MLP, and action-level embedding.

Evaluation on the challenging Breakfast \cite{kuehne2014language}, YouTube Instructional \cite{alayrac2016unsupervised}, and 50Salads \cite{stein2013combining} datasets demonstrate our superior performance over the state of the art.

The rest of the paper is organized as follows:  Sec.~\ref{sec:Related Work} reviews related work, Sec.~\ref{sec:Self-Supervised Learning} presents our SSL, Sec.~\ref{sec:Models} specifies our HMM and its inference, Sec.~\ref{sec:Learning} formalizes our joint training, Sec.~\ref{sec:AllActivities} extends our approach to a more general setting, and Sec.~\ref{sec:Experiments} presents our experiments.


\section{Related Work}\label{sec:Related Work}
This section reviews closely related work on action segmentation under a reduced level of supervision.  

{\bf Transcript supervised action segmentation}. In this problem, we have access to the true action ordering in training videos, but their exact action boundaries are unknown. While this problem is different from ours, our approach draws motivation from recent advances. For example, Extended Connectionist Temporal Classification (ECTC) regularizes action alignment with consistency of frame similarity \cite{huang2016connectionist}.  Some approaches alternatively align actions with frames and update their action models \cite{kuehne2017weakly, richard2017weakly}. Other methods first estimate a video segmentation, and then use this estimation to train a classifier for frame labeling \cite{richard2018neuralnetwork,li2019weakly}. 

{\bf Set supervised action segmentation}. In this problem, we have access to the ground-truth set of actions occurring in the training video, but we do not know their ordering and temporal extents.  This problem was first addressed with multi-instance learning \cite{richard2018action}. In \cite{li2020set}, Set-Constrained Viterbi (SCV) algorithm is used to produce pseudo-ground-truth labels of frames for the subsequent fully supervised action segmentation. In \cite{Fayyaz_2020_CVPR}, Set Constrained Temporal Transformation (SCT) is used to predict action labels of oversegmented temporal regions.

{\bf Unsupervised action segmentation and detection} focuses on exploiting the temporal structure of videos \cite{SaxenaCVPR15,   vidalmata2020joint, fernando2017unsupervised, sener2018unsupervised, Jain18}. For example, \cite{fernando2017unsupervised}  detects all pairs of matching video segments, \cite{SaxenaCVPR15} learns co-occurrence and temporal relations between actions, \cite{sener2018unsupervised} proposes a Generalized Mallows Model to jointly learn action appearance and temporal structure, \cite{Jain18} groups consecutive frames to form communities similar to social networks. 
Sec.~\ref{sec:Intro} summarizes \cite{kukleva2019unsupervised} as the most closely related approach to ours, and points out our differences and extensions. 

{\bf Temporal SSL} has been recently used in various video interpretation problems, including action segmentation, for learning a temporal feature embedding. Existing temporal SSL methods are typically aimed at capturing the temporal structure of video either at the frame level or at the entire video level. Examples of frame-level temporal SSL include the following: \cite{misra2016shuffle} determines whether a sequence of frames from a video is shuffled or in the correct temporal order; \cite{LeeHS017} predicts a permutation of a sequence of frames; and \cite{Ommer18, Kim_Cho_Kweon_2019} estimate both spatial and temporal ordering of frame patches. Also, examples of video-level temporal SSL include the following: \cite{JenniECCV20}  learns a feature embedding that captures changes in the video's motion dynamics such as speed and warping; \cite{wei2018learning} identifies the arrow of time in videos; and \cite{EpsteinCV20, BenaimELMFRID20,Yao_2020_CVPR} predict the playback speed.

All of these approaches perform the data manipulation for their temporal SSL without taking into account the ordering of action segments in videos. For example, when they shuffle frames in a given video \cite{misra2016shuffle} or when they manipulate the video's speed \cite{BenaimELMFRID20, JenniECCV20}, the correct frame ordering or speed is readily available and deterministic, given by the very input video. Instead, our video manipulation shuffles latent actions whose ``correct'' ordering is only inferred and not necessarily well-aligned with ground truth (since the ground truth is not available in unsupervised learning). Also, our SSL accounts for higher-level temporal structures in the video beyond a sequence of frames.

\section{Action Shuffle SSL}
\label{sec:Self-Supervised Learning}
\begin{figure}[b]
\centering
\includegraphics[width=\linewidth]{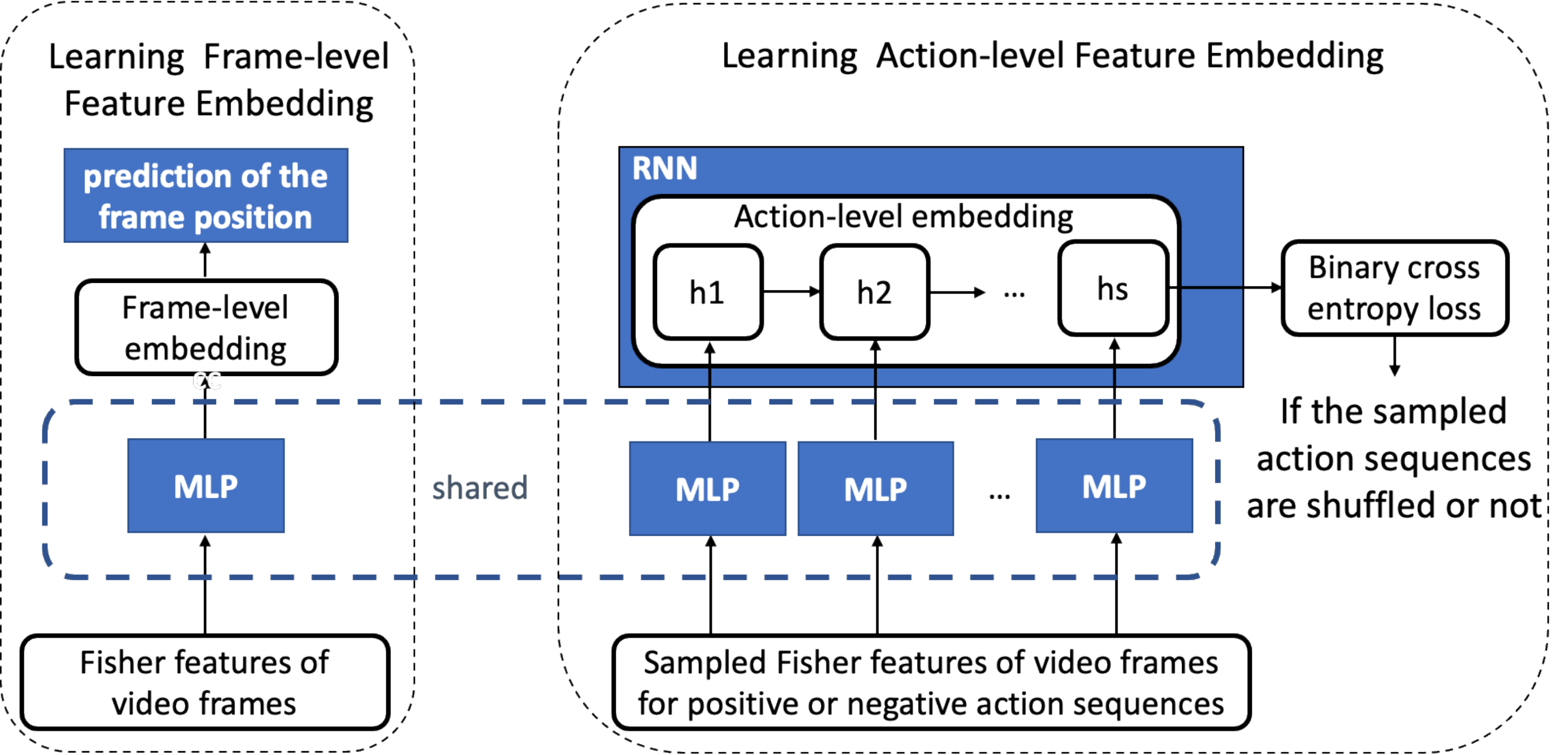}
\caption{(Left) Learning the frame-level embedding of \cite{kukleva2019unsupervised}. (Right) Learning our action-level embedding. The sampled positive and negative action sequences are input to an RNN for predicting if they are shuffled or not. The MLP of both frame-level embedding and our action-level embedding is shared.}
\label{fig:Embedding}
\end{figure}

Our temporal feature embedding rests on the assumption that actions tend to occur in similar relative locations across videos showing the same activity. We capture this temporal consistency at the level of frames and the level of actions. As shown in Fig.~\ref{fig:Embedding}, we first learn the frame-level temporal embedding as in \cite{kukleva2019unsupervised}, in the initial stage of our approach (see Fig~\ref{fig:Overview}a), and then iteratively learn our action-level embedding through the Generalized EM algorithm  (see Fig~\ref{fig:Overview}b). In each iteration of the Generalized EM, our action-level embedding is updated as described below.

Fisher features of video frames of positive and negative action sequences are input to an RNN for predicting whether the input actions are shuffled. Both positive and negative sequences consist of 3 action segments randomly sampled from a video, where every action segment has 5 consecutive frames randomly selected from that action's time interval in the video. The sampled actions are shuffled in negative sequences, and respect their time ordering in positive sequences. Since the frame-level embedding is learned as the output of an MLP, as in \cite{kukleva2019unsupervised}, our SSL {\em shares the same} MLP to produce the input to our RNN. We train the RNN on the binary cross entropy loss, and take its hidden layer as our action-level feature embedding.

\section{HMM and Its Inference}
\label{sec:Models}

Videos are represented as sequences of frame features $\bm{x}_{1:T_m} = [x_1,...,x_t,...,x_{T_m}]$, where $T_m$ is the length of $m$th video, $m=1,\dots,M$. For HMM modeling and inference, $x_t$ represents the action-level feature embedding.   Following  \cite{kukleva2019unsupervised,alayrac2016unsupervised,sener2018unsupervised}, we assume that there are at most $N$ latent actions, $\mathbb{C}=\{c:c=1,\dots,N\}$, and that each $c$ may occur only once in a  video. Thus, for a given video $\bm{x}_{1:T}$, our goal is to find an optimal action segmentation, $(\hat{\bm{c}}_{1:K}, \hat{\bm{l}}_{1:K})$, where $\hat{\bm{c}}_{1:K}=[\hat{c}_1,\dots,\hat{c}_{K}]$ is the predicted action sequence, $K\le N$, $\hat{c}_k\in\mathbb{C}$, $k$ is the index of a video segment $k=1,\dots,K$, and $\hat{\bm{l}}_{1:K} = [\hat{l}_1,\dots,\hat{l}_{K}]$ are their corresponding lengths such that $\sum_{k=1}^K\hat{l}_k=T$.

{\bf The HMM.}  We specify the following HMM for identifying a MAP action segmentation: %
\begin{equation}
\begin{array}{l}
p(\bm{c}_{1:K}, \bm{l}_{1:K}|\bm{x}_{1:T})   \\ \quad  \propto p(\bm{x}_{1:T}|\bm{c}_{1:K}, \bm{l}_{1:K})p(\bm{l}_{1:K}|\bm{c}_{1:K})p(\bm{c}_{1:K}) , \\ \quad 
   = \displaystyle \Big[\prod_{t=1}^{T} p(x_t|c_{k(t)})\Big] \Big[\prod_{k=1}^{K}  p(l_k| c_k) \Big] \Big[\prod_{k=1}^{K-1}  p(c_{k+1}| c_k) \Big].
   \end{array}
   \label{eq:Markov}
\end{equation}
In \eqref{eq:Markov}, the likelihood $p(x_t|c)$ is estimated as
\begin{equation}
    p(x_t|c) \propto \frac{p(c|x_t)}{p(c)},
    \label{eq:framelikelihood}
\end{equation}
where $p(c|x_t)$ is computed by passing $x_t$ to the MLP and taking the resulting softmax score for action $c$. The prior $p(c)$ is defined as a uniform distribution for all $c\in\mathbb{C}$. The length of each action is modeled as a Poisson distribution:
\begin{equation}
    p(l|c) = \frac{\lambda_{c}^{l} }{l!}e^{-\lambda_{c}} , 
    \label{eq:Poisson}
\end{equation}
where $\lambda_{c}$ is the mean length for class $c$. 

The transition probability $p({c_{k+1}}|c_k)$ in \eqref{eq:Markov} is defined with respect to the initialized ordering of the latent actions, $\bm{c}^{\text{init}}_{1:N}=[1,2,3,\dots,N]$, computed by the K-means in the initial stage of our unsupervised learning, as mentioned in Sec.~\ref{sec:Intro}. Specifically, we define:
\begin{equation}
p({c_{k+1}}|c_k) \propto 
\left\{\begin{array}{lcl}\displaystyle
     \frac{\lambda_{c_{k}}+\lambda_{c_{k+1}}}{\sum_{j=c_k}^{c_{k+1}}\lambda_{c_j}}&,& c_{k+1}>c_k \\
         0& ,&\text{otherwise}
    \end{array} \right.,
    \label{eq:transition}
\end{equation}
where ``$\propto$'' denotes the appropriate normalization such that $\sum_{c=c_k+1}^N p(c|c_k) = 1$. From \eqref{eq:transition}, $p({c_{k+1}}|c_k)$ prevents transitions that would result in the opposite action ordering from the initial $\bm{c}^{\text{init}}_{1:N}$. Also, $p({c_{k+1}}|c_k)$ favors transitions between actions with consecutive labels in $\bm{c}^{\text{init}}_{1:N}$ the most. But we also allow  transitions to actions with larger labels, $(c_{k+1}-c_k)> 1$, especially if the expected time interval between $c_k$ and $c_{k+1}$  is short, along the initial action ordering in $\bm{c}^{\text{init}}_{1:N}$. This allows for skipping some of the latent actions in the HMM inference. Importantly,  $p({c_{k+1}}|c_k)$ penalizes ``long skips'' since the denominator in \eqref{eq:transition} strictly increases with the number of actions skipped $(c_{k+1}{-}c_k{-}1)$.

Note that our $p({c_{k+1}}|c_k)$ extends previous work \cite{kukleva2019unsupervised}, where $p({c_{k+1}}|c_k)=0$ for all $(c_{k+1}-c_k)\ne 1$, i.e., in \cite{kukleva2019unsupervised}, the predicted action sequence is made equal to the fixed initial transcript, $\hat{\bm{c}}_{1:N}=\bm{c}^{\text{init}}_{1:N}$. Another extension of \cite{kukleva2019unsupervised} is our length model $p(l|c)$ in  \eqref{eq:Poisson}.

{\bf The HMM Inference.} The MAP $(\hat{\bm{c}}_{1:K}, \hat{\bm{l}}_{1:K})$ is computed by the Viterbi inference. Given  $\bm{x}_{1:T}$ and the initial transcript $\bm{c}_{1:N}^{\text{init}}$, the Viterbi algorithm recursively maximizes the posterior in \eqref{eq:Markov} such that the first $k$ actions of the transcript $\bm{c}_{1:k}=[c_1,...,c_k]$ at time $t$ respect the action ordering in  $\bm{c}_{1:N}^{\text{init}}$, $\bm{c}_{1:k} \preceq \bm{c}_{1:N}^{\text{init}}$
\begin{equation}
\begin{array}{l}
\displaystyle
  p(\hat{\bm{c}}_{1:k},\hat{\bm{l}}_{1:k}|\bm{x}_{1:t}) = \left.\max_{\substack{t',\; t'<t\\c_k\in\bm{c}_{k-1:N}^{\text{init}}}}\right\{ p(\hat{\bm{c}}_{1:k-1},\hat{\bm{l}}_{1:k-1}|\bm{x}_{1:t'})\\ 
  \displaystyle \quad \quad   \cdot \left.\left(\prod_{s = t'}^{t} p(x_s|c_{k(s)})\right) \cdot p(l_k|c_k) \cdot p(c_{k}|c_{k-1})\right\},
  \end{array}
\label{eq:Viterbi}
\end{equation}
where $l_k = t-t'$ and $p(c_{k}|c_{k-1})$ penalizes transitions for which $c_k - c_{k-1}>1$. We set $p(\cdot|\bm{x}_{1:0}) = 1$, and $p(c_1|c_0) = \kappa$, where $\kappa>0$ is a constant. The final prediction is given by the final recursion of $p(\hat{\bm{c}}_{1:K},\hat{\bm{l}}_{1:K}|\bm{x}_{1:T})$.

{\bf The Likelihood MLP.} Our HMM uses an MLP to estimate frame likelihoods, as specified in \eqref{eq:framelikelihood}. Note that this likelihood MLP is different from the MLP used in our SSL and shown in Fig.~\ref{fig:Embedding}. The likelihood MLP is a framewise classifier, initially trained on annotations produced by the K-means in the initial stage of our approach (see Fig.~\ref{fig:Overview}a). In the subsequent training iterations, the likelihood MLP is supervised by the frame labelings, $\{\hat{\bm{c}}_{1:K}\}_{m=1}^M$, predicted in the previous iteration for all videos.

\section{Alternating Learning}
\label{sec:Learning}

We jointly learn the HMM, MLP, and action-level feature embedding by alternating the Expectation and Maximization steps of the Generalized EM algorithm.  In the E-step, we estimate the following $Q (\theta,\theta^{\text{old}})$ function of the $\theta$ parameters, where $\theta = \{W, \Lambda\}$, $W$ are the parameters of the likelihood MLP and RNN for the action-level embedding,  and $\Lambda=\{\lambda_c:c\in\mathbb{C}\}$ is the set of mean action lengths: %
\begin{equation}
\begin{array}{l}
Q(\theta, \theta^{\text{old}}) =\\ \frac{1}{T_m} \sum_{({\bm{c}}^m,\bm{l}^m)} \sum_{m=1}^M p(\bm{c}^{m},{\bm{l}}^{m}|\bm{x}^{m};\theta^{\text{old}}) \log p(\bm{c}^{m},{\bm{l}}^{m},\bm{x}^{m};\theta),
\end{array}
\label{eq:Theoretical EM}
  \end{equation}  
where from \eqref{eq:Markov} the joint log-likelihood in \eqref{eq:Theoretical EM} is
\begin{equation}
 \begin{array}{l} 
 \log p(\bm{c}^{m},{\bm{l}}^{m},\bm{x}^{m};\theta)=\\
  \quad \sum_{t=1}^{T_m}\log p(x_t^{m}|c_{k(t)}^{m};W) +  \sum_{k=1}^{K_m} \log  p(l_k^{m}| c_k^m;\Lambda)\\
  \quad + \sum_{k=1}^{K_m}\log p(c_{k+1}^{m}| c_k^m;\Lambda)
\end{array}
\label{eq:loglikelihood}
\end{equation}
For our videos, the posteriors are products of framewise likelihoods over typically more than 1000 frames. Hence, the posteriors in \eqref{eq:Theoretical EM} are very close to zero for all latent variables $({\bm{c}}^m,\bm{l}^m)$ that are different from the MAP ones, $({\bm{c}}^m,\bm{l}^m)\ne (\hat{{\bm{c}}}^m,\hat{\bm{l}}^m)$, specified in \eqref{eq:Viterbi}. Therefore, for our problem setting, it is appropriate to approximate \eqref{eq:Theoretical EM} as $Q(\theta, \theta^{\text{old}}) \approx  \frac{1}{T_m} \log p(\hat{\bm{c}}^{m},\hat{\bm{l}}^{m},\bm{x}^{m};\theta) $.

In  the M-step, we maximize $Q(\theta, \theta^{\text{old}})$ with respect to $W$ and $\Lambda$ by performing fixed-step gradient descent which updates the parameters of the likelihood MLP and RNN as:
\begin{align}
W^{\text{new}} = W^{\text{old}} + \alpha \frac{1}{M} \sum_{m=1}^M\sum_{t=1}^{T_m}\nabla \log p(x_t^{m}|\hat{c}_{k(t)}^{m};W),
\end{align}
where $\alpha$ is the learning rate. Also, maximizing  $Q(\theta, \theta^{\text{old}})$ with respecto $\Lambda$ gives the following update rule for the mean length of every action $\lambda_c\in\Lambda$:
\begin{align}
   \lambda_c^{\text{new}} =\lambda_c^{\text{old}} + \frac{1}{M}\sum_{m=1}^M\left(\frac{\sum_{k=1}^{{K}_m} \hat{l}_{k}^{m}\cdot 1(c=\hat{c}_{k}^{m})}{\sum_{k=1}^{{K}} 1(c=\hat{c}_{k}^{m})} - \lambda_c^{\text{old}}\right)
\end{align}
where $M$ is the number of videos. 

In \cite{tanner2012tools}, the interested reader can find the proof that the weak convergence is guaranteed, $p(\bm{c}_{1:k},{\bm{l}}_{1:k}|\bm{x}_{1:t};\theta^{\text{new}}) > p(\bm{c}_{1:k},{\bm{l}}_{1:k}|\bm{x}_{1:t};\theta^{\text{old}})$, as long as the above updating process maintains that $Q(\theta^{\text{new}}, \theta^{\text{old}}) > Q(\theta^{\text{old}}, \theta^{\text{old}})$. In our experiments, we empirically observe convergence when a difference between the new and old log-posteriors is less than $\epsilon=10^{-3}$.

\section{Learning with All Activities}
\label{sec:AllActivities}
As in \cite{kukleva2019unsupervised}, we also address action segmentation across all activities. In this case, we have no access to the activity class, so the assumption that actions tend to occur in similar relative temporal orderings across all videos may not be justified. To address this issue, we take the following steps. 

Similar to the previously discussed one-activity setting, we first learn the frame-level embedding for all videos across all activities. The frames are then clustered in the frame-level embedding space to construct a bag-of-words representation for each video with a soft assignment. This allows for clustering videos based on their bag-of-words representation, and each resulting cluster is interpreted as a set of videos showing the same activity. Finally, we perform the second stage of our approach, as illustrated in Fig.~\ref{fig:Overview}b, for each video set separately.

\section{Experiments}\label{sec:Experiments}

{\bf Datasets.}  For evaluation, we use three benchmark datasets: Breakfast \cite{kuehne2014language}, YouTube Instructional \cite{alayrac2016unsupervised}, and 50Salads \cite{stein2013combining}.  

Breakfast is a large-scale dataset, consisting of 10 different complex activities of people making breakfast, with approximately 8 actions per activity class. Every video has on average 6.9 actions, and the video lengths vary from a few seconds to several minutes. For evaluation on Breakfast, as unsupervised input video features, we use the reduced Fisher Vector features \cite{kuehne2016end}, as in \cite{sener2018unsupervised,kukleva2019unsupervised}.

YouTube Instructions shows five activities: \textit{making coffee, cpr, jumping car, changing car tire, potting a plant} in 150 videos with an average  length of about two minutes. As in \cite{sener2018unsupervised,kukleva2019unsupervised}, for evaluation on YouTube Instructions, we use the unsupervised features proposed by \cite{alayrac2016unsupervised}.

50Salads has 4.5 hours of video footage of one complex activity, that of making a salad. Its video length is much longer than that of Breakfast and YouTube Instructions. As in \cite{kukleva2019unsupervised}, two different action-granularity levels are used for evaluation: mid-level with 17 action classes and eval-level with 9 action classes.

{\bf Evaluation Metrics.} For establishing a correspondence between the predicted segmentation and ground-truth segments, we follow \cite{alayrac2016unsupervised, sener2018unsupervised, kukleva2019unsupervised} and use the Hungarian algorithm for one-to-one matching based on the overlap between the matched segments over all videos.
For evaluation on Breakfast and 50Salads, we compute the mean over frames (MoF). For YouTube Instructions, we report the F1-score, where for calculating precision and recall the positive detections must overlap more than 50\% with the matched ground-truth segments. 

{\bf Implementation Details.} In our experiments, the dimension of our action-level feature embedding is set to be 20. The MLP used for the embedding has one $40\times 20$ hidden layer. The hidden-to-hidden layer in the RNN is $20 \times 20$. The likelihood MLP for the HMM has one $40 \times N$ hidden layer.  One iteration of the Generalized EM algorithm is referred to as epoch. In our experiments, we observe convergence after 20 epochs. It only takes several hours for training, \eg an average of 1 hour training time given an activity from Breakfast. We observe as the number of epochs increases we get non-decreasing $Q$ function given by \eqref{eq:Theoretical EM}. Within each epoch we generate $2\cdot M$ positive and negative action sequences, two per video, for our SSL of the RNN.  The backprop for the SSL is performed with the SGD with momentum, and our learning rate is 0.001.

{\bf Ablations.} We consider the following variants of our approach for evaluating the effect of each component:
\begin{itemize} [itemsep=0pt,topsep=0pt, partopsep=0pt]
\item \abbrapproach{} = Our full approach with the alternating learning of the action-level embedding and HMM;
\item FTE + HMM = From our full approach we removed the action-level embedding and keep the frame-level temporal embedding (FTE) of \cite{kukleva2019unsupervised}; the HMM is iteratively updated, but not FTE as it belongs to the first stage of our approach;
\item ActionShuffle + initHMM = From our full approach we remove the alternating learning, but use the action-level embedding and the HMM initialized on the K-means results from the first stage of our approach; this version tests the effect of our joint learning.
\item ActionShuffle + Viterbi = From our full approach we remove the HMM, but use the action-level embedding and the Viterbi algorithm constrained to respect the initial fixed transcript of actions as in \cite{kukleva2019unsupervised}; this version amounts to running \cite{kukleva2019unsupervised} with our action-level feature embedding instead of their FTE.
\end{itemize}

\subsection{Evaluation for the Same Activity}
In this section, evaluation is done for the setting where all videos belong to the same activity class.  Table~\ref{Table:breakfast result}, Table~\ref{Table:YouTube Instructions result}, and Table~\ref{Table:50salads result} show that our approach outperforms the state of the art, on all three datasets,  for this setting. On Breakfast, our approach gives a higher F-1 score by 11.5, and a higher MoF by 10.7 than the strong baseline \cite{kukleva2019unsupervised}. On Youtube Instructional, our approach outperforms \cite{kukleva2019unsupervised} by 3.8 in F1-score and 5.9 in MoF. On 50salads, we get better results than \cite{kukleva2019unsupervised} for both eval and mid granularity level features by 3.7 and 4.2 in MoF, respectively. In addition, for a comparison to an upper-bound performance, Table~\ref{Table:breakfast result} also includes the best results of recent approaches on Breakfast trained under two other learning settings which increase the level of supervision -- specifically, fully-supervised and weakly transcript-supervised learning as reviewed in Sec.~\ref{sec:Related Work}. As can be seen, our approach comes in performance very close to the best weakly supervised approach on Breakfast. In addition, we also show the results of LSTM+AL \cite{aakur2019perceptual}, as a representative of approaches that use a different evaluation method with a per-video local Hungarian matching.

\begin{table}[ht]
\begin{center}
\begin{tabular}{l c c}
\thickhline 
& {\bf Breakfast}  &\\
\hline
Fully Supervised & & MoF \\
\hline
HTK \cite{kuehne2014language} & & 28.8 \\
TCFPN \cite{ding2018weakly} & &  52.0\\
HTK+DTF w. PCA \cite{kuehne2016end}& &  56.3\\
RNN+HMM \cite{ding2018weakly} & & 60.6\\
\hline
\hline
Weakly Supervised & &  MoF\\
\hline
OCDC \cite{bojanowski2014weakly} &  & 8.9  \\
HTK \cite{kuehne2017weakly} &  & 25.9 \\
CTC \cite{huang2016connectionist} &   & 21.8 \\
ECTC \cite{huang2016connectionist} &  & 27.7 \\
HMM+RNN \cite{richard2017weakly} &   & 33.3 \\
TCFPN \cite{ding2018weakly} &   & 18.3 \\
NN-Viterbi \cite{richard2018neuralnetwork} &  & 43.0 \\
D3TW \cite{chang2019d3tw} & & 45.7 \\
CDFL \cite{li2019weakly} &  & 50.2 \\
\hline
\hline
Unsupervised & F1-score & MoF \\
\hline
Mallow \cite{sener2018unsupervised}  & - & 34.6 \\
CTE \cite{kukleva2019unsupervised}   & 26.4 & 41.8 \\
(LSTM+AL) \cite{aakur2019perceptual} & - & ($42.9^{*}$) \\
VTE-UNET \cite{vidalmata2021joint} & - & 48.1 \\
{\bf Our \abbrapproach{}} & {\bf 37.9} & {\bf 52.5}\\
\thickhline
\end{tabular}
\end{center}
\caption{Comparison of our approach with the state of the art on Breakfast. The table also shows the latest best results on Breakfast for the fully supervised and weakly supervised learning, as an upper-bound performance to ours.
The dash means ``not reported'', and $^*$ means that results are evaluated with the ``per video'' Hungarian matching, not the Hungarian matching over all videos. In the unsupervised setting, we get the best F1-score and MoF, and come very close to the best weakly supervised performer.}
\label{Table:breakfast result}
\end{table}

\begin{table}[ht]
\begin{center}
\begin{tabular}{l c c}
\thickhline 
& {\bf YouTube Instructions}  &\\
\hline
Unsupervised &  & \\
\hline
 & F1-score & MoF \\
\hline
Frank-Wolfe \cite{alayrac2016unsupervised} & 24.4 & - \\
Mallow \cite{sener2018unsupervised}  & 27.0 & 27.8 \\
CTE \cite{kukleva2019unsupervised}   & 28.3 & 39.0 \\
VTE-UNET \cite{vidalmata2021joint} & 29.9 & - \\
{\bf Our \abbrapproach{}} & {\bf 32.1} & {\bf 44.9}\\
(LSTM+AL) \cite{aakur2019perceptual} & ($39.7^*$) & - \\
\thickhline
\end{tabular}
\end{center}
\caption{Comparison of our approach with the state of the art under unsupervised learning on YouTube Instructions. The dash means ``not reported'', and $^*$ means that results are evaluated with the ``per video'' Hungarian matching, not the Hungarian matching over all videos. We achieve the best F1-score and MoF.}
\label{Table:YouTube Instructions result}
\end{table}

\begin{table}[ht]
\begin{center}
\begin{tabular}{l c c}
\thickhline 
& {\bf 50Salads}  &\\
\hline
Unsupervised && \\
\hline
  & Granularity level & MoF \\
\hline
VTE-UNET \cite{vidalmata2021joint} & eval & 30.6 \\
CTE \cite{kukleva2019unsupervised} &eval &35.5\\
{\bf Our \abbrapproach{}} &eval & {\bf 39.2}\\
(LSTM+AL) \cite{aakur2019perceptual} & eval & ($60.6^*$) \\
\hline
VTE-UNET \cite{vidalmata2021joint} & mid & 24.2 \\
CTE \cite{kukleva2019unsupervised} &mid &30.2\\
{\bf Our \abbrapproach{}} &mid & {\bf 34.4}\\
\thickhline
\end{tabular}
\end{center}
\caption{Comparison of our approach with the state of the art under unsupervised learning on 50salads. The dash means ``not reported'', and $^*$ means that results are evaluated with the ``per video'' Hungarian matching, not the Hungarian matching over all videos. We achieve the best results on both action-granularity levels.}
\label{Table:50salads result}
\end{table}

\begin{figure}
\centering
\includegraphics[width=\linewidth]{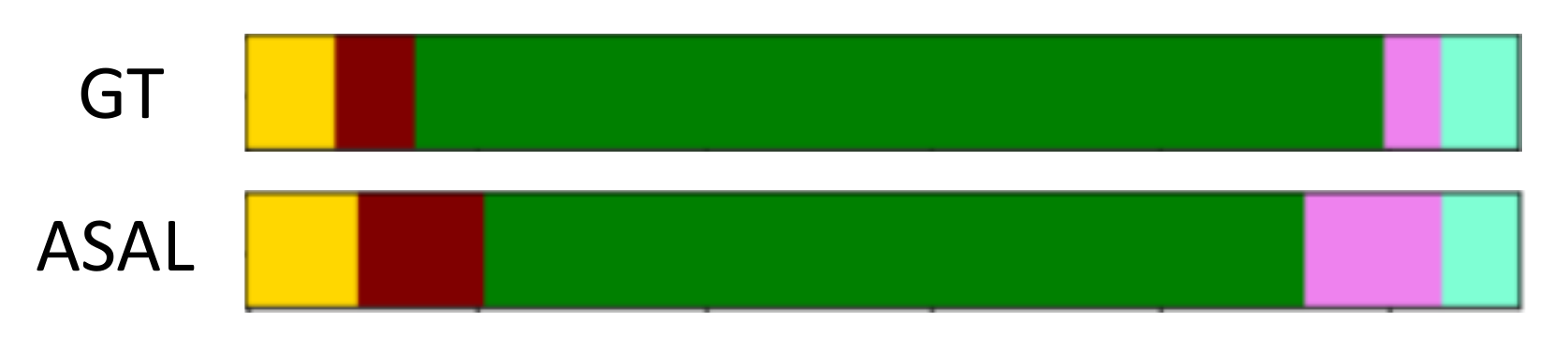}
\caption{A representative result for our full approach ASAL on a sample video \textit{P04\_webcam02\_P04\_friedegg} from the Breakfast dataset. Top-down, the rows correspond to the ground truth sequence of actions (\textcolor{mygold}{pour\_oil}, \textcolor{mymaroon}{crack\_egg},  \textcolor{mygreen}{fry\_egg}, \textcolor{mypink}{take\_plate}, \textcolor{myaquamarine}{put\_egg2plate}) and our action segmentation.}
\label{fig:result}
\end{figure}

Figure~\ref{fig:result} illustrates a representative action segmentation that our approach ASAL produced for a sample video from Breakfast. As can be seen, ASAL is usually good at identifying salient actions, but may miss the start and end frames of the corresponding ground-truth actions. 

{\bf Effect of Feature Embedding.}
For the setting where all videos belong to the same activity class,  we compare the results of our full approach \abbrapproach{} and FTE+HMM. FTE+HMM  computes the frame-level temporal embedding (FTE) in the first stage, and does not update it in the second stage. Table.~\ref{Table:Effect of Embedding} shows that, on Breakfast, our full approach \abbrapproach{} gives better results than FTE+HMM by 2.1 in F1-score and 4.0 in MoF. The large performance gain of our approach  suggests that our action-level embedding successfully captures the temporal structure of videos.

\begin{table}[ht]
\begin{center}
\begin{tabular}{l c c}
\thickhline 
&{\bf Embeddings}  &\\
\hline
Breakfast && \\
\hline
  & F1-score & MoF \\
\hline
FTE + HMM & 35.4 & 47.7 \\
{\bf \abbrapproach{}} &{\bf 37.9} & {\bf 52.5}\\
\thickhline
\end{tabular}
\end{center}
\caption{Evaluation of different feature embeddings on Breakfast. ASAL is our full approach, and FTE+HMM does not use our action-level embedding but only the frame-level embedding of \cite{kukleva2019unsupervised}.
We achieve better results, which suggests that our action-level embedding successfully captures the temporal structure.}
\label{Table:Effect of Embedding}
\end{table}

{\bf Effect of Alternating Training and HMM.}
For the setting where all videos belong to the same activity class, we evaluate two different ways for learning  the components of our approach -- specifically, the joint alternating learning of our full approach  \abbrapproach{}, and separate learning of the action-level embedding and HMM in ActionShuffle+initHMM based on the K-means results. Specifically, ActionShuffle+initHMM trains the likelihood MLP with the framewise pseudo-ground truth from the K-means, and learns the expected action length in \eqref{eq:Poisson} as an average of action lengths in the K-means. Then, such an HMM is inferred with the Viterbi algorithm, and the resulting action segmentation is used for our SSL of the action-level feature embedding. In ActionShuffle+initHMM, there is no alternating training, i.e., after the initial separate learning all components are not updated further.  In addition, we also evaluate ActionShuffle+Viterbi that does not have our HMM but the model used in \cite{kukleva2019unsupervised}. This model evaluates frame likelihoods, but does not capture action lengths and action transitions. Also, their model inference is constrained to the fixed initial transcript of actions. Table.~\ref{Table:HMM Training} shows that our full approach gives superior performance on Breakfast in comparison with ActionShuffle+initHMM and ActionShuffle+Viterbi. ActionShuffle+initHMM  gives better results than ActionShuffle+Viterbi, which suggests that accounting for action lengths and transitions in the HMM is very important.

\begin{figure}
\centering
\includegraphics[width=\linewidth]{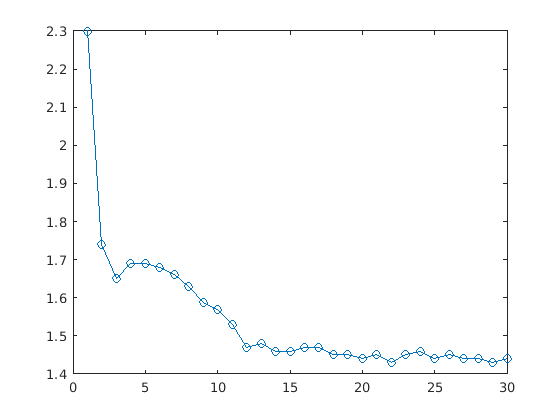}
\caption{The mean of Q values over all videos for training epochs on videos of the activity ``Juice" from Breakfast. The plot shows that our Generalized EM algorithm converges after 20th epoch.}
\label{fig:Q_value}
\end{figure}

For evaluating convergence of our alternating learning, in Fig.~\ref{fig:Q_value}, we plot values of the Q function, given by \eqref{eq:Theoretical EM}, over training epoches on the activity ``Juice" from Breakfast. The vertical axis is the mean of Q values over all videos in one epoch. The figure shows that our Generalized EM algorithm converges after 20th epoch on the activity ``Juice".

\begin{table}[ht]
\begin{center}
\begin{tabular}{l c c}
\thickhline 
{\bf Difference in Learning and Models} &  &\\
\hline
Breakfast & F1-score & MoF\\
\hline
ActionShuffle+Viterbi & 30.7 & 44.1 \\
ActionShuffle+initHMM & 32.1 & 46.8 \\
{\bf \abbrapproach{}} &{\bf 37.9} & {\bf 52.5}\\
\thickhline
\end{tabular}
\end{center}
\caption{Effect of Alternating Training and HMM. We compare our full approach \abbrapproach{} with ActionShuffle+Viterbi, where no HMM is used, and ActionShuffle+initHMM, where no joint training is used. \abbrapproach{} gives the best performance.}
\label{Table:HMM Training}
\end{table}

\subsection{Evaluation across All Activities}
In this section, we evaluate our approach in a more general unsupervised setting, where videos belong to different activity classes with different temporal structures of actions. Following  \cite{kukleva2019unsupervised}, we also perform the Hungarian matching of all predicted actions to the ground truth action segments. For this setting, all experiments are evaluated on the Breakfast dataset. We assume $N = 5$ actions for each activity, and that there are 10 activities in Breakfast. Then the matching is performed between 50 different action clusters to 48 ground-truth actions, and the remaining unmatched clusters are set as background. 

In Table~\ref{Table:Learning across All Activities}, we compare our full approach \abbrapproach{} with the state of the art on Breakfast, for this multi-activity setting, and observe that \abbrapproach{} gives the best results.
 
\begin{table}[ht]
\begin{center}
\begin{tabular}{l c }
\thickhline 
{\bf Learning across All Activities}  &\\
\hline
Breakfast   & MoF  
\\
\hline
CTE \cite{kukleva2019unsupervised} & 16.4 \\
{\bf Our \abbrapproach{}} & {\bf 20.2}\\
\thickhline
\end{tabular}
\end{center}
\caption{Evaluation in the setting when videos belong to multiple different activity classes. Our full approach \abbrapproach{} gives better results than \cite{kukleva2019unsupervised}.}
\label{Table:Learning across All Activities}
\end{table}

{\bf Effect of Embedding}
For the multi-activity setting, we compare our \abbrapproach{} with FTE+HMM which does not use the action-level feature embedding. We also define another variant of our approach, where the action-level embedding is learned for all activity classes without taking into account that videos belong to different activities -- the approach referred to as Global emb+HMM. As shown in Table~\ref{Table:Effect of Embedding across all activities}, the action-level embedding learned for each estimated activity in \abbrapproach{} gives better performance than Global emb+HMM, since actions of different activities have very different temporal orderings which cannot be reliably captured by our the action-level embedding aimed only for a single activity.

\begin{table}[ht]
\begin{center}
\begin{tabular}{l c }
\thickhline 
{\bf Embedding Across All Activities}  &\\
\hline
Breakfast   & MoF 
 \\
\hline
Global emb + HMM & 18.3 \\
FTE + HMM  & 17.9 \\
{\bf \abbrapproach{}} & {\bf 20.2}\\
\thickhline
\end{tabular}
\end{center}
\caption{Evaluation of embeddings across all activities. The superior performance of our \abbrapproach{} suggests that the action-level embedding is not suitable for capturing large variations in the temporal structure of multiple distinct activities.}
\label{Table:Effect of Embedding across all activities}
\end{table}

{\bf Effect of Alternating Learning and HMM}
For the multi-activity setting, we also compare our full approach \abbrapproach{} with ActionShuffle+initHMM and ActionShuffle+Viterbi. As shown in Table~\ref{Table:Effect of HMM Training}, ASAL gives the superior performance.

\begin{table}[ht]
\begin{center}
\begin{tabular}{l c }
\thickhline 
{\bf HMM \& Training Across All Activities}  &\\
\hline
Breakfast  & MoF 
 \\
\hline
ActionShuffle+Viterbi & 17.2 \\
ActionShuffle+initHMM & 18.7 \\
{\bf \abbrapproach{}} & {\bf 20.2}\\
\thickhline
\end{tabular}
\end{center}
\caption{Evaluation of HMM and two different learning strategies on Breakfast for the setting when videos belong to different activities. We compare our full approach \abbrapproach{} with ActionShuffle+Viterbi, where no HMM is used, and ActionShuffle+initHMM, where learning of the HMM and action-level embedding is independent and not alternated.  \abbrapproach{} gives the best performance.}
\label{Table:Effect of HMM Training}
\end{table}

\section{Conclusion}
In this paper, we have advanced unsupervised action segmentation by making the following contributions. First, we have specified a new self-supervised learning (SSL) as a verification of the temporal ordering of actions. The proposed SSL provides the action-level feature embedding used in an HMM for inferring a MAP action segmentation. Second, our HMM explicitly models action lengths and transitions, and in this way relaxes the restrictive assumptions of prior work that all videos have a fixed time ordering of actions. Third, we have unified learning of the action-level embedding and HMM within the Generalized EM framework. Our evaluation studies two different settings -- when videos show a single activity or multiple distinct activities -- on the Breakfast, Youtube Instructions, and 50Salads datasets. In both unsupervised settings, and on all three datasets, our approach achieves superior results relative to the state of the art, and even comes close to the best weakly supervised performer on Breakfast for the single-activity setting. We have also presented a detailed ablation study that demonstrates advantages of the proposed: a) action-level embedding relative to the frame-level temporal embedding of prior work; b) modeling of action lengths and transitions relative to fixing inference to a predefined action transcript as in prior work; and c) joint alternating training of all components of our approach relative to their separate training.
\\

\noindent{\bf{Acknowledgement}}. This work was supported in part by DARPA XAI Award N66001-17-2-4029 and DARPA MCS Award N66001-19-2-4035.

{\small
\bibliographystyle{ieee_fullname}
\bibliography{egbib}
}

\end{document}